%% file: egpaper_final.tex
\documentclass[10pt,twocolumn,letterpaper]{article}

\usepackage{cvpr}
\usepackage{times}
\usepackage{epsfig}
\usepackage{graphicx}
\usepackage{amsmath}
\usepackage{amssymb}
\usepackage{subcaption}
\usepackage{overpic}
\usepackage{multirow}
\usepackage{makecell}
\usepackage{cancel}

\newcommand{\class}{y}
\newcommand{\attri}{z}
\newcommand{\data}{x}
\newcommand{\features}{f}

\newcommand{\idxb}{k}

\newcommand{\numattri}{e}

\newcommand{\mref}{\texttt{M-REF}}
\newcommand{\mfi}{\texttt{M-FI}}
\newcommand{\miacd}{\texttt{M-IACD}}
\newcommand{\mdacd}{\texttt{M-DACD}}
\newcommand{\mcdia}{\texttt{M-CDIA}}
\newcommand{\mcdda}{\texttt{M-CDDA}}

\newcommand{\val}{h}
\newcommand{\classVec}{Y}

\usepackage[breaklinks=true,bookmarks=false]{hyperref}

\DeclareMathOperator*{\argmax}{arg\,max}

\cvprfinalcopy 

\def\cvprPaperID{****} 

\begin{document}

\title{Dependency Decomposition and a Reject Option for Explainable Models}

\author{Jan Kronenberger\\
Computer Science Institute\\
Ruhr West University of Applied Sciences\\
{\tt\small jan.kronenberger@hs-ruhrwest.de}
\and
Anselm Haselhoff\\
Computer Science Institute\\
Ruhr West University of Applied Sciences\\
{\tt\small anselm.haselhoff@hs-ruhrwest.de}
}

\maketitle

\input{abstract.tex}

\input{into.tex}
\input{state.tex}
\input{method_save.tex}
\input{results.tex}
\input{conclusion.tex}

{\small
\bibliographystyle{ieee}
\bibliography{References}
}

\end{document}

%% file: abstract.tex
\begin{abstract}
Deploying machine learning models in safety-related domains (e.g. autonomous driving, medical diagnosis) demands for  approaches that are explainable, robust against adversarial attacks and aware of the model uncertainty. Recent deep learning models perform extremely well in various inference tasks, but the black-box nature of these approaches leads to a weakness regarding the three requirements mentioned above. Recent advances offer methods to visualize features, describe attribution of the input (e.g. heatmaps), provide textual explanations or reduce dimensionality. However, are explanations for classification tasks dependent or are they independent of each other? For instance, is the shape of an object dependent on the color? What is the effect of using the predicted class for generating explanations and vice versa? 

In the context of explainable deep learning models, we present the first analysis of dependencies regarding the probability distribution over the desired image classification outputs and the explaining variables (e.g. attributes, texts, heatmaps).  Therefore, we perform an Explanation Dependency Decomposition (EDD). We analyze the implications of the different dependencies and propose two ways of generating the explanation. Finally, we use the explanation to verify (accept or reject) the prediction.
\end{abstract}

%% file: into.tex
\section{Introduction}

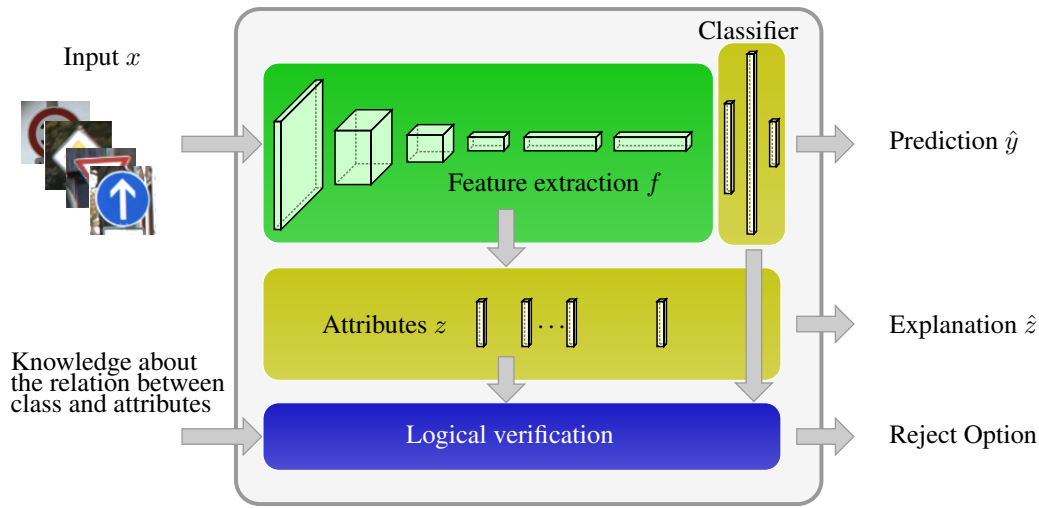
\begin{figure*}[ht]
	\centering
	\begin{overpic}[width=0.8\linewidth, tics=2]
		{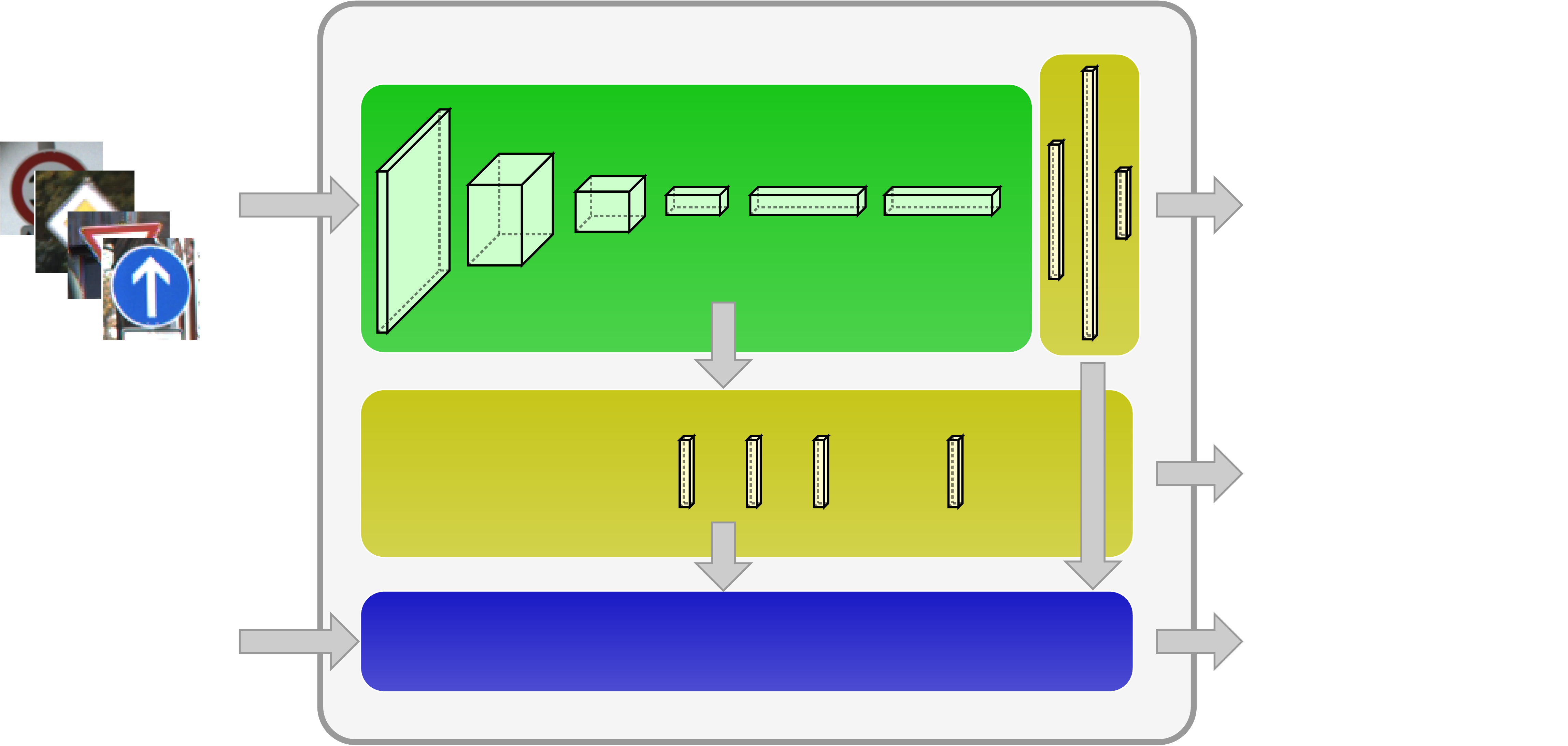}
		\put(28,16.5){ Attributes $\attri$}
		\put(48.5,17){ \dots}
		\put(4,42){Input $\data$}
		\put(82,34){ Prediction $\hat{\class}$}
		\put(82,16.5){ Explanation $\hat{\attri}$}
		\put(82,6){ Reject Option}
		\put(40,30){ Feature extraction $\features$}
		\put(63.8,44.5){ Classifier}
		\put(36,6){ \textcolor{white}{Logical verification}}
		\put(-1,13){Knowledge about}
		\put(-1,11){the relation between}
		\put(-1,9){class and attributes}
	\end{overpic}
	\caption{Overview of the approach. In addition to a base-network that performs the intrinsic task (e.g. classifying traffic signs), an auxiliary model is added to explain the decision. The CNN extracts features for the classification of traffic signs and the explaining attributes. The main contribution of this work is an analysis of different dependencies between the explaining attributes and the classifier. Finally, the explaining attributes are used to verify the decision (predicted class).}
	\label{fig:approach}
\end{figure*}
In recent years the success of deep neural networks was mainly driven by technological advances in network architectures and learning algorithms to improve the performance of the model. However, using these models in real-world applications an interaction with humans is inevitable. This interaction leads to new considerations regarding ethical and legal issues as well as aspects of user acceptance. Especially in safety-critical applications the models need to be explainable. Therefore, decisions, predictions or recommendations of the machine learning model must be comprehensible for the user. For example, an autonomous vehicle detecting a stop-sign or executing a full braking maneuver should justify its decision. Predictions of classical methods like decision trees that can be explained by introspection and temporal fusion methods (e.g. Kalman filtering) are often enriched with an uncertainty measure (error covariance). In contrast, deep learning models are often treated as black-box models without any explanation. Even  probabilistic outputs of neural networks, which could be treated as a confidence measurement, are often poorly calibrated \cite{Guo2017} as the confidence estimate must match the accuracy. Recently, Kendall and Gall \cite{Gal2017} proposed Bayesian deep learning to model the needed uncertainty.

Understanding the internal representation that neural networks are constructing is one step into the direction of understanding their way of reasoning and thus reduce the chance of malfunctions or successful attacks. Early work on hierarchical representations were already presented by Hubel and Wiesel in 1962 \cite{Hubel1962}. Their approach of complex cells, which are activated by simpler cells, leads to an internal representation of differently complex objects and properties within the network. These internal representations can be used to justify decisions and make them comprehensible \cite{Xu2018}. They can also help to develop decision-making systems that are not biased, unfair or even racist \cite{Thelisson2017}. 

In this work we focus on methods that provide explanations in form of attribution of the input (e.g. visual concepts) or textual explanations. Fig.~\ref{fig:approach} shows an example, where an auxiliary network is used for explaining the prediction besides the base-network that performs the intrinsic task (e.g. classifying traffic signs). The explanation could be a sentence containing the attributes (color, shape, symbols, etc.) that have contributed to the decision process or a region in the input space that was most important for the prediction. The appearance of attributes can also be used to verify the prediction by using their joint probability and optionally reject the decision.

First, we introduce a loss function to jointly learn explaining variables (attributes) and classification outputs. The loss can be mapped to a broad class of explainable models (e.g. \cite{Hendricks2016,Ribeiro2016}). Second, we decompose or the factorize the dependencies of attributes and classification outputs in different ways, to analyze the relations between explanations and the intrinsic task. In addition, appropriate network architectures and a method for attribute-based verification including a reject option are introduced.
The evaluation is based on the german traffic sign recognition benchmark (GTSRB, \cite{Stallkamp2012}).

%% file: state.tex
\section{Related Work}

Convolutional neural networks already create internal representations as proposed by \cite{Alsallakh2018,Zhou2014}. In order to make these representations usable for the human user, they can be synthesized in different ways. Previous approaches solve this problem by visualization of features \cite{Bach2015,Dumitru2009,Lapuschkin2016,Nguyen2017,Nguyen2014,olah2017feature,Ribeiro2016,Selvaraju2017,Zeiler2014,Zeiler2011}, attribution \cite{Fong2017,Hand2017,Huang2016,Kindermans2017,Kindermans2017a,Simonyan2013,Springenberg2014,Sundararajan2017,Zhang2018} or by descriptive sentences \cite{Hendricks2016,Vinyals2015,Xu2015}. However, these approaches have several disadvantages. Visualizations of activations and filters may lead to a comprehensive explanation of the internal representations. However, they are difficult to understand by the user because the visualizations are of high dimensionality. Image caption systems are often more likely to generate generic captions and in some cases are not tailored to give explanations for a specific decision \cite{Lindh2018}. We adress theis problem by using general as well as specific attributes. Another alternative is to use semantic parts to justify decisions \cite{Zhang2018}. These semantic parts describe small objects or visual concepts and are assigned to the individual classes by a voting system. Finding a meaningful description of a class can be very difficult, since the differences between individual examples within a class can be very large, while different classes are sometimes very similar \cite{He2017}. To create meaningful synthetic training data for such applications, more extensive methods are required~\cite{Peng2018}.

Choosing the attributes to explain the prediction is a hard task as it requires prior knowledge. While we have chosen very simple handcrafted attributes in our work like in other approaches \cite{Zhang2018}, it is also possible to have them selected unsupervised by the network \cite{Huang2016}. However, this high variety of attributes might not be usable for easy explanations as labeling the unsupervised attributes is difficult due to the possibility of false interpretation.
 
Having comprehensible decision-making systems is very important because simple predictions systems can be fooled quite easily \cite{Akhtar2017,Nguyen2014} by modifying the input. These manipulations can be either modifications of the real object (e.g. contamination on traffic signs, graffiti,~\dots) or adversarial attacks which are designed to confuse the DNNs by adding a small amount of noise \cite{Co2019,Goodfellow2015,Kurakin2017}. However, while it is already possible to prevent certain adversarial attacks by denoising the input data \cite{Liao2018} modifications on the real objects are difficult to recognize because the network compensates such errors and still predicts the most likely class. With our method such unclear inputs can be detected and rejected if necessary attributes are missing.

%% file: method_save.tex
\section{Methods}
In this paper we pose the supervised learning problem in terms of a probability distribution $p(\class|\data; \theta )$, where $\class$ denotes the class in the set of all classes $\mathcal{Y}$, $\data$ denotes the input data or image and $\theta$ is the parameter vector of the network. In explainable models we are interested in jointly learning the distribution of the output class and the explaining variables $\attri$; we call these variables attributes.
The joint probability distribution of the class and attributes given the input data can be written as 
\begin{equation}
p(\class, \attri|\data; \theta).
\end{equation}
In general the parameter vector $\theta$ is determined using maximum likelihood estimation. We use a cross-entropy loss between the empirical data distribution $\hat{p}_{data}$ and the model distribution $p$ given by
\begin{equation}
	L(\theta)= -\mathbb{E}_{\class, \attri, \data \sim \hat{p}_{data}} \log p(\class, \attri|\data; \theta).
\end{equation}

\subsection{Explanation Dependency Decomposition (EDD)}
This section presents the details of the proposed method regarding the decomposition of different dependencies between attributes and class-labels. We start with a reference model, where no explaining variables are included. Based on this reference, the dependencies are integrated into the probabilistic model and the assumptions as well as the implications are discussed. 
In the following sections we distinguish between parameters of the feature extractor $\theta_{f}$ (convolutional layers), classifier $\theta_{\class}$ and explanation or attributes $\theta_{\attri}$, respectively. Up to now we have used the parameter vector $\theta = \theta_{\class,\attri, f}$ to denote all model parameters, that is $\theta_{\class,\attri, f}=(\theta_{\class},\; \theta_{\attri},\; \theta_{f})$. The joint probability $p(\class, \attri |\data; \theta)$ can be decomposed as either 
\begin{align}
	p(\class, \attri|\data; \theta) &= p(\class|\data; \theta) p(\attri | \class, \data; \theta)\text{, or}
	\label{equ:no6}\\
		p(\class, \attri|\data; \theta) &= p(\class|\attri, \data; \theta) p(\attri | \data; \theta).
	\label{equ:no5}
\end{align}
Variable $\class$ is depended on $\attri$ or vice versa. $p(\attri | \class, \data; \theta)$ from Eq.~\ref{equ:no6} can be decomposed into the single attributes contained in $\attri$
\begin{equation}
	p(\attri | \class, \data; \theta) = p(\attri_{\numattri}| \class, \data; \theta) \prod_{\idxb = 1}^{\numattri - 1} p(\attri_{\idxb}|\text{pa}\left({\attri_{\idxb}}\right), \class, \data; \theta).
\end{equation}
Eq.~\ref{equ:no5} can be decomposed in a similar fashion. In this context $\text{pa}(\attri_{\idxb})$ describes the parental variables of $\attri_{\idxb}$ \cite{Barber2012}. Depending on the model, we used either Eq.~\ref{equ:no6} or Eq.~\ref{equ:no5} with different parental variables. Tab.~\ref{tab:models} summarizes the dependencies and the used equations for all utilized models.
\begin{table}[!ht]
	\centering
	\caption{Overview of the dependencies of the different models.}
	\begin{tabular}{|c|c|c|}
		\hline 
		Model & Base Equation & $\text{pa}\left({\attri_{\idxb}}\right)$\\ 
		\hline 
		\mref & $p(\class|\data; \theta)$ & n/a \\ 
		\mfi &  $p(\class|\data; \theta)p(\attri|\data; \theta)$  & $\emptyset$ \\ 
		\miacd & Eq.~\ref{equ:no6} & $\class$ \\
		\mdacd & Eq.~\ref{equ:no6} & $\{\attri_{\idxb+1},\dots , \attri_{\numattri} \}$\\
		\mcdia & Eq.~\ref{equ:no5} & $\emptyset$\\
		\mcdda & Eq.~\ref{equ:no5} &  $\{\attri_{\idxb+1},\dots , \attri_{\numattri}\}$\\
		\hline 
	\end{tabular}
	\label{tab:models}
\end{table}
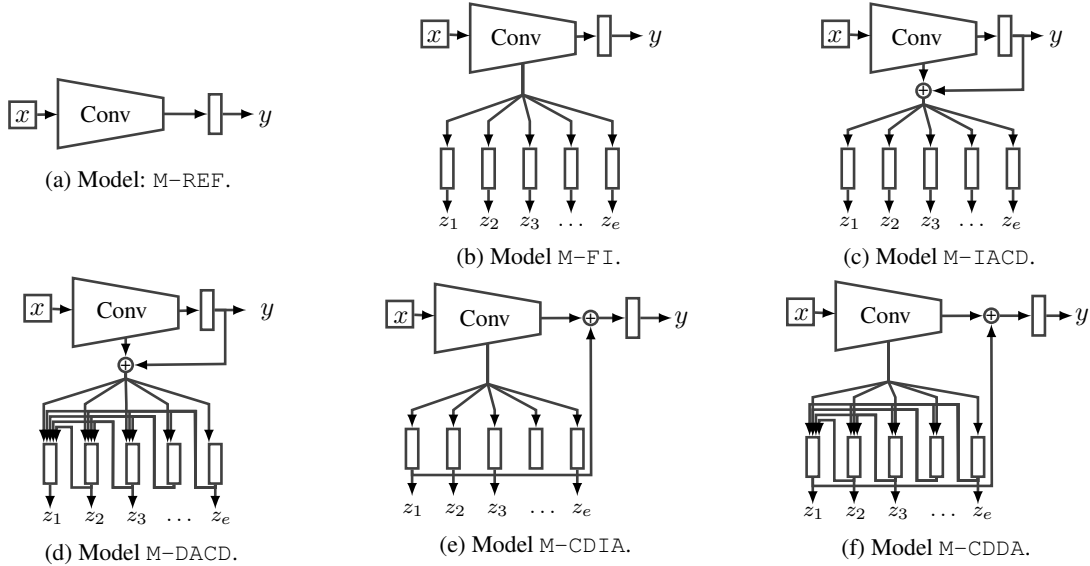
\begin{figure*}
	\centering
	\begin{subfigure}[c]{0.3\textwidth}
		\centering
		\begin{overpic}[scale=.7, tics=10]
			{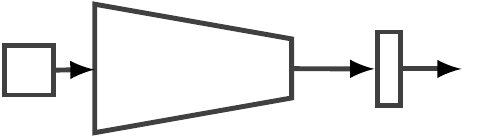}
			\put(28,11){\small Conv}
			\put(3,11){$x$}
			\put(96,11){$y$}
		\end{overpic}
		\caption{Model: \mref.}
		\label{fig:mref}
	\end{subfigure}
	\begin{subfigure}[c]{0.3\textwidth}
		\centering
		\begin{overpic}[scale=.7, tics=10]
			{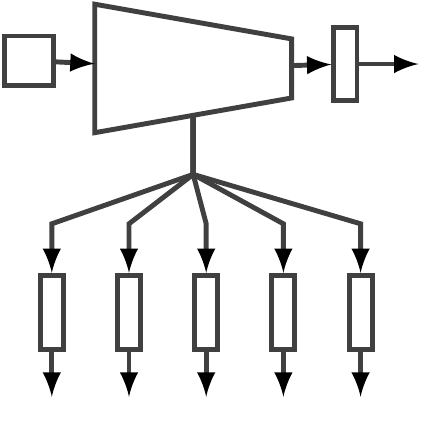}
			\put(30,77){\small Conv}
			\put(3,78){$x$}
			\put(96,78){$y$}
			\put(8,1){\footnotesize $z_1$}
			\put(25,1){\footnotesize $z_2$}
			\put(42,1){\footnotesize $z_3$}
			\put(58,1){\footnotesize $\dots$}
			\put(76,1){\footnotesize $z_e$}
		\end{overpic}
		\caption{Model \mfi.}
		\label{fig:mfi}
	\end{subfigure}
	\begin{subfigure}[c]{0.3\textwidth}
		\centering
		\begin{overpic}[scale=.7, tics=10]
			{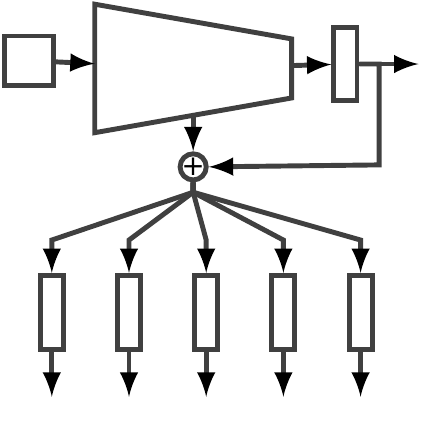}
			\put(30,77){\small Conv}
			\put(3,78){$x$}
			\put(96,78){$y$}
			\put(8,1){\footnotesize $z_1$}
			\put(25,1){\footnotesize $z_2$}
			\put(42,1){\footnotesize $z_3$}
			\put(58,1){\footnotesize $\dots$}
			\put(76,1){\footnotesize $z_e$}
		\end{overpic}
		\caption{Model \miacd.}
		\label{fig:miacd}
	\end{subfigure}
	\begin{subfigure}[c]{0.3\textwidth}
		\centering
		\begin{overpic}[scale=.7, tics=10]
			{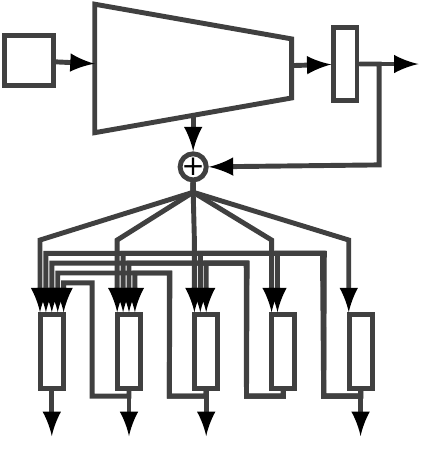}
			\put(30,83){\small Conv}
			\put(3,84){$x$}
			\put(96,84){$y$}
			\put(8,1){\footnotesize $z_1$}
			\put(25,1){\footnotesize $z_2$}
			\put(42,1){\footnotesize $z_3$}
			\put(58,1){\footnotesize $\dots$}
			\put(76,1){\footnotesize $z_e$}
		\end{overpic}
		\caption{Model \mdacd.}
		\label{fig:mdacd}
	\end{subfigure}%
	\begin{subfigure}[c]{0.3\textwidth}
		\centering
		\begin{overpic}[scale=.7, tics=10]
			{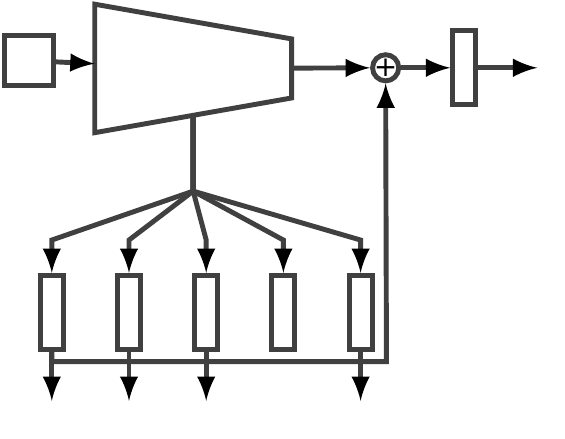}
			\put(25,63){\small Conv}
			\put(3,64){$x$}
			\put(96,64){$y$}
			\put(6,1){\footnotesize $z_1$}
			\put(20,1){\footnotesize $z_2$}
			\put(34,1){\footnotesize $z_3$}
			\put(48,1){\footnotesize $\dots$}
			\put(62,1){\footnotesize $z_e$}
		\end{overpic}
		\caption{Model \mcdia.}
		\label{fig:mcdia}
	\end{subfigure}
	\begin{subfigure}[c]{0.3\textwidth}
		\centering
		\begin{overpic}[scale=.7, tics=10]
			{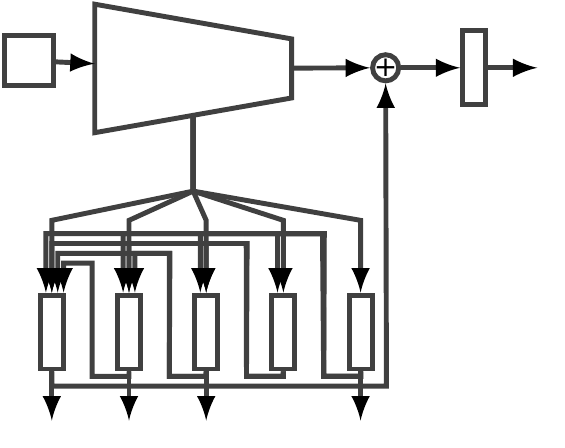}
			\put(25,65){\small Conv}
			\put(3,66){$x$}
			\put(96,66){$y$}
			\put(6,1){\footnotesize $z_1$}
			\put(20,1){\footnotesize $z_2$}
			\put(34,1){\footnotesize $z_3$}
			\put(48,1){\footnotesize $\dots$}
			\put(62,1){\footnotesize $z_e$}
		\end{overpic}
		\caption{Model \mcdda.}
		\label{fig:mcdda}
	\end{subfigure}
	\caption{Overview of the used models. While \mref \ is used to measure the influence of the additional attributes to the pristine classifier $\class$ the other models differ according to their dependencies. Model \mfi \ has independent classifiers. The models \miacd \ and \mdacd \ have a dependency of the attributes $\attri$ on the class $\class$. With model \mdacd \, the attributes are also independent. In the models \mcdia \ and \mcdda \ the dependencies are reversed. For the sake of clarity, the individual links between the convolution layers and the attribute classifiers have been merged.}
	\label{fig:models}
\end{figure*}
As a reference model \mref\ we use a standard classification model, without any explaining variables. This model is used to verify the proposition of \cite{Turner2015} that additional attributes used for explanation may limit the learning freedom of the network and thus lead to worse results. Fig.~\ref{fig:mref} shows the schematic representation of the network divided into the input data $\data$, the convolution network, the classifier (fully connected layer), and the class-output $\class$.

\subsubsection{Full Independence (\mfi)}
The simplest way of integrating explanations is to assume full independence. The model \mfi\ was developed under the assumption that the attributes are independent of each other in the same way in which the class is independent of the attributes. The implication of these assumptions are, that, given the input data $\data$ the attributes $\attri$ are not providing any additional information to solve the classification task $\class$ and vice versa. 
It is assumed, that during the joint training process the attributes and classes can have an influence on the parameter adjustment of feature extractor $\theta_f$. In contrast, the parameters of the model that are dedicated to the attributes $\theta_{\attri}$ are not adjusted by the classification loss and vise versa. 
The corresponding network structure is visualized in Fig.~\ref{fig:mfi}.

\subsubsection{Attributes with a Class-Dependency (\miacd, \mdacd)}

When designing an explainable model we want to get an explanation for a specific decision of the model. In this case the assumption that the explaining attributes are independent of the class may be oversimplified. To incorporate the class information we use the model \miacd, where we only assume independent attributes.
Thus, the explaining part of the model is given the ability to focus on class-specific explanations. 
The joint training process of the class and attributes may be unstable, since the attribute model has to use the noisy class outputs. To overcome this problem we apply a teacher forcing using the ground-truth labels during the first iterations of the training process. The network structure is shown in Fig.~\ref{fig:miacd} and it is obvious that the explaining part of the model can influence all parameters of network, including the classifier. This aspect is possibly restricting the classification performance.  

Using \miacd \, as a base-model, thus preserving the class-dependency of the attributes, in addition a dependency among each attribute can be considered (cp. Fig.~\ref{fig:mdacd}). This model has the minimal amount of assumptions possible; as well as model \mcdda. As an example, the model can capture the dependence of object shape and color attributes. 

\subsubsection{Class with a Attribute-Dependency (\mcdia, \mcdda)}
In the previous models we have assumed a specific order while decomposing the joint probability distribution $p(\class, \attri |\data; \theta)$. The goal was to get an explanation for a specific decision of a classifier. A different way to look at the problem is to give a possible explanation first (what kind of attributes are visible in the input space?) and define a classifier to leverage explanations for an enhanced classification performance. Thus, we still obtain an explainable model, but with slight shift of the objective with a focus on the classification performance. 
This type of model with a classifier dependent on the attributes and independence among attributes is denoted as model \mcdia. The corresponding network structure is given in Fig. \ref{fig:mcdia}.

The last decomposition of $p(\class, \attri |\data; \theta)$ is based on model \mcdia, but without making any assumptions. Likewise, a dependence of the class on the attributes is used and in addition the dependence among the attributes is preserved.
The structure of model \mcdda \, is shown in Fig. \ref{fig:mcdda}.

\subsection{Explanation}
Our models represent their explanation by the presence and absence of attributes. The explanation made by the DNN can be seen as an image specific explanation 

\begin{equation}
	\hat{\class}, \hat{\attri} = \argmax_{\class, \attri} p({\class, \attri | \data; \theta})
\end{equation}
as they explain the resulting decision with the attributes visible in the image. The output of the DNN $\left(\hat{\class}, \hat{\attri}\right)$ is the most plausible combination of a class and the associated explanation. The explanation doesn't need any knowledge about the internal dependencies, but may be affected by them.   

\subsection{Reject Option}
The predicted attributes $\hat{\attri}$ can not only be utilized for explaining the decision process, but also to verify, support or reject a prediction of the base-network. In our application we can define sufficient and necessary conditions (denoted as $C$) for each class based on the attributes. \textit{Necessary} attributes must be recognized when a class is predicted. Furthermore, we can directly deduce a class if \textit{sufficient} attributes of exclusively that class are recognized. For example given only a detected \textit{bicycle} symbol we can directly induce class \textit{Bicycle lane}. In general we can define at least some necessary conditions that could be used to support the decision. A simple categorization of the outputs for attribute-based verification of a prediction $\hat{\class}$ is given by Belnap~\cite{Belnap1977}, who has introduced the categories
\begin{enumerate}
	\item \textit{True} - we only have information about $\hat{\class}$ being true (no information about $\hat{\class}$ being false). 
	\item \textit{Both} - we have information about $\hat{\class}$ being true or false (uncertain).
	\item \textit{False} - we only have information about $\hat{\class}$ being false.
	\item \textit{None} - we have no information.
\end{enumerate}
Depending on the application at hand we could utilize different Belnap categories to define a reject option. We use the Belnap category \textit{True} as a strong condition to accept a prediction and the three other categories define a reject. 
In order to apply the Belnap categories we use the predicted attributes $\hat{\attri}$ to define a possible set of class $\hat{\classVec}$. The set includes all classes that meet the defined conditions $C$.
The reject option or verification can then be expressed by
\begin{align*}
{\val}\left(\hat{\class}, \hat{\classVec}\right)=\left\{ \begin{array}{cc}
accept,&  \text{ if } \hat{\class} \in \hat{\classVec}  \land \hat{\classVec}\setminus \hat{\class}=\emptyset \\ 
reject,&  \text{otherwise}
\end{array}.   \right.
\end{align*}

\begin{table*}[ht!]
	\centering
	\caption{The accuracy of class predictions and explanations is evaluated for the different models. In addition the reject option based on the attributes is evaluated only on the accepted predictions. The rejection rate defines the number of samples with uncertain predictions (no decision possible).}
	\begin{tabular}{|l|r|r|r|r|r|r|}
		\hline
	Model & \mref & \mdacd & \miacd & \mfi & \mcdda & \mcdia \\ \hline
		\hline
	\multicolumn{7}{|c|}{Evaluation of the class prediction and explanation} \\
	\hline
		Accuracy of $\hat{\class}$  & $92.20$ & $92.03$  & $90.75$ & $92.34$& $97.86$  & \boldmath $98.04$ \\ 
		Accuracy of $\hat{\attri}$  & & \boldmath $89.85$  & $87.36$ & $84.95$& $89.78$  & $88.28$ \\ 
		\hline
		\multicolumn{7}{|c|}{Evaluation of the class prediction with reject option} \\
		\hline
		Accuracy of $\hat{\class}$ & & $95.29$  & $95.33$  & $98.14$ & $99.27$ & \boldmath $99.35$  \\ 
		Rejection rate & & \boldmath $10.15$ & $12.64$ & $15.05$  & $10.22$  & $11.72$  \\ 
		\hline
	\end{tabular}
	\label{tab:acc}
\end{table*}

\section{Experiments}


\subsection{Dataset}

We use the dataset GTSRB~\cite{Stallkamp2012} for our experiments. It consists of $43$ different classes showing german traffic signs. The a priori distribution of the classes are compensated by augmentation. Furthermore, the data is normalized to have zero mean and a standard deviation of one, to compensate for different lighting conditions and a bias in exposure. For the classes we have determined the following attributes sorted by complexity:
\begin{itemize}
	\item \textit{Simple}: Main and border color (white, red, blue, black, yellow)
	\item \textit{Medium}: Shapes (round, triangular, square and octagonal)
	\item \textit{Complex}: Numbers (0, \dots, 9) and symbols (car, truck, stop, animal, ice, children, people, construction site, attention, traffic lights, bicycle, narrow point, uneven)
\end{itemize}
\input{figures/gtsrbsynt.tex}
In order to cover different variations of attributes, synthetic data, as shown in Fig.~\ref{fig:gtsrbsynt}, is included. 
These synthetic samples account for approximately $39~\%$ of the dataset. Due to the selected attributes, it is possible that several attributes are only available for a single class (for example, the sign "priority road" is the only one with a square shape and a yellow main color). In order to prevent the respective classifiers from accidentally learning the wrong property, the synthetic data is used to create square signs with different main colors.

\subsection{Architecture}

The architecture of the convolution network is based on AlexNet~\cite{Krizhevsky2012} with a reduced number of parameters and an input size of $48 \times 48$. Instead of the original $6.2\cdot 10^7$ parameters in the AlexNet, the model~\mref \ uses only $2.9\cdot 10^5$ parameter. The remaining networks need between $4.6\cdot 10^6$ and $4.8\cdot 10^6$ parameter, depending on the complexity of the dependencies. The reason for fewer parameters is mainly a parameter reduction  in the fully-connected layers. As the internal representations become more complicated with the depth of the network \cite{Riesenhuber1999,Zeiler2014}, the attributes are classified based on different feature layers of the network. Simple attributes, such as the main or border color, are determined directly after the second convolution layer. After the third layer the shapes are predicted. The remaining complex attributes (symbols, numbers) are determined after the fourth layer.

%% file: figures/gtsrbsynt.tex
\begin{figure}[!ht]
	\centering
	\begin{subfigure}[b]{0.09\linewidth}
		\centering
		\includegraphics[width=0.75\linewidth]{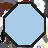} 
	\end{subfigure}
	\begin{subfigure}[b]{0.09\linewidth}
		\centering
		\includegraphics[width=0.75\linewidth]{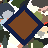} 
	\end{subfigure}
	\begin{subfigure}[b]{0.09\linewidth}
		\centering
		\includegraphics[width=0.75\linewidth]{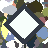} 
	\end{subfigure}
	\begin{subfigure}[b]{0.09\linewidth}
		\centering
		\includegraphics[width=0.75\linewidth]{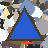}
	\end{subfigure}
	\begin{subfigure}[b]{0.09\linewidth}
		\centering
		\includegraphics[width=0.75\linewidth]{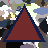} 
	\end{subfigure}
	\begin{subfigure}[b]{0.09\linewidth}
		\centering
		\includegraphics[width=0.75\linewidth]{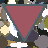} 
	\end{subfigure}
	\begin{subfigure}[b]{0.09\linewidth}
		\centering
		\includegraphics[width=0.75\linewidth]{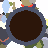} 
	\end{subfigure}
	\begin{subfigure}[b]{0.09\linewidth}
		\centering
		\includegraphics[width=0.75\linewidth]{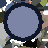}
	\end{subfigure}
	\begin{subfigure}[b]{0.09\linewidth}
		\centering
		\includegraphics[width=0.75\linewidth]{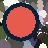} 
	\end{subfigure}

	\begin{subfigure}[b]{0.09\linewidth}
		\centering
		\includegraphics[width=0.75\linewidth]{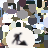} 
	\end{subfigure}
	\begin{subfigure}[b]{0.09\linewidth}
		\centering
		\includegraphics[width=0.75\linewidth]{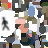} 
	\end{subfigure}
	\begin{subfigure}[b]{0.09\linewidth}
		\centering
		\includegraphics[width=0.75\linewidth]{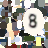} 
	\end{subfigure}
	\begin{subfigure}[b]{0.09\linewidth}
		\centering
		\includegraphics[width=0.75\linewidth]{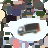}
	\end{subfigure}
	\begin{subfigure}[b]{0.09\linewidth}
		\centering
		\includegraphics[width=0.75\linewidth]{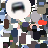} 
	\end{subfigure}
	\begin{subfigure}[b]{0.09\linewidth}
		\centering
		\includegraphics[width=0.75\linewidth]{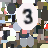} 
	\end{subfigure}
	\begin{subfigure}[b]{0.09\linewidth}
		\centering
		\includegraphics[width=0.75\linewidth]{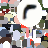} 
	\end{subfigure}
	\begin{subfigure}[b]{0.09\linewidth}
		\centering
		\includegraphics[width=0.75\linewidth]{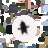}
	\end{subfigure}
	\begin{subfigure}[b]{0.09\linewidth}
		\centering
		\includegraphics[width=0.75\linewidth]{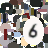} 
	\end{subfigure}
	\caption{Some examples of synthetic data. The colors, shapes and symbols are sampled from the original dataset.}
	\label{fig:gtsrbsynt}
\end{figure}

%% file: results.tex
\subsection{Results}

\subsubsection{Class predictions and explanations}

The experiments show, that adding an auxiliary model for explaining a DNN doesn't have a significantly negative effect on the accuracy. The models have to be compared to the pristine classifier \mref \ that has an accuracy of $92.20~\%$ (cp. Tab.~\ref{tab:acc}). However, the predictions of the additional attributes can have a very positive effect on improving the prediction of the class. Model \mcdda\ and \mcdia\ are examples that use a classifier with a dependency on the explanation and therefore can improve the classification performance. This suggests that the prediction of the class benefits from the additional knowledge about the attributes and the additional parameters available in the network. The accuracy of the explanations of these two models is close to the optimal model \mdacd.
In contrast, the models that focus on class specific explanations (\mdacd\ and \miacd) provide less accurate class predictions. This property may be due to the fact that the attributes depend on the class and therefore their training has an influence on the classifier network. However, as expected the best performance regarding the explanation is achieved (\mdacd). 
Finally, using full independence of classes and explanations doesn't change the performance of the classifier compared to the reference model. The explanations delivered by this kind of model are not of comparable quality given by all other models. The results for all models are presented in the first two rows of Tab.~\ref{tab:acc}.    
\subsubsection{Class predictions with reject option}

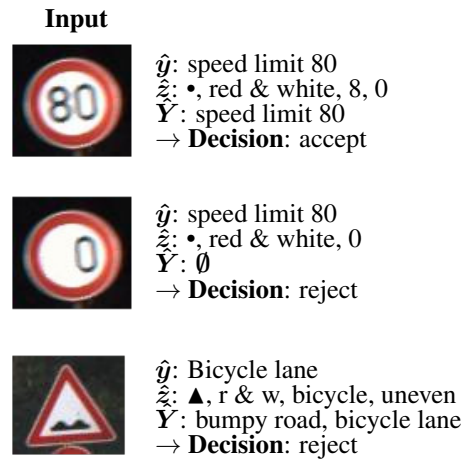
\begin{figure}[ht]
	\centering
	\begin{overpic}[width=.8\linewidth, tics=10]
		{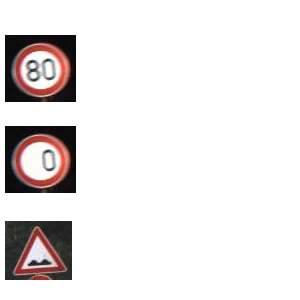}
		\put(8,92){\textbf{Input}}
		\put(30,83){\boldmath $\hat{\class}$: speed limit 80}
		\put(30,78){\boldmath $\hat{\attri}$: \textbullet, red \& white, 8, 0}
		\put(30,73){\boldmath $\hat{\classVec}$: speed limit 80}
		\put(30,68){\textbf{$\rightarrow$ Decision}: accept}
		\put(30,53){\boldmath $\hat{\class}$: speed limit 80}
		\put(30,48){\boldmath $\hat{\attri}$: \textbullet, red \& white, 0}
		\put(30,43){\boldmath $\hat{\classVec}$: $\emptyset$}
		\put(30,38){\textbf{$\rightarrow$ Decision}: reject}
		\put(30,22){\boldmath $\hat{\class}$: Bicycle lane}
		\put(30,17){\boldmath $\hat{\attri}$: $\blacktriangle$, r \& w, bicycle, uneven}
		\put(30,12){\boldmath $\hat{\classVec}$: bumpy road, bicycle lane}
		\put(30, 7){\textbf{$\rightarrow$ Decision}: reject}
	\end{overpic}
	\caption{Examples of three decisions using the reject option.}
	\label{fig:verification}
\end{figure}

The second part of Tab.~\ref{tab:acc} describes the accuracy of the predictions with the option of rejecting a decision. The accuracy for all models increases while between $10.15\%$ and $15.05\%$ of the decisions are rejected. The model \mfi\ has the highest rejection rate because the classifier and the explaining model are fully independent. Therefore, it is likely that the explanation and the class prediction deliver contrary results. However, the accuracy of class prediction increases by almost $6\%$ using the reject option. The best accuracy with respect to the class prediction is again obtained with model \mcdia. This model has a moderate rejection rate of $11.72\%$ compared to the lowest rejection rate of $10.15\%$ (\mdacd). 
It is worth to mention that this kind of reject option justifies an acceptance or rejection. In Fig.~\ref{fig:verification} three different outcomes of the verification are shown. In the first example the decision of the classifier and the explanation agree. Therefore, the decision of the classifier can be accepted. There are multiple reasons for getting a reject. In the second example one \textit{necessary} attribute ($8$) that is required for the class is missing. In this case $\hat{Y}$ is empty (category \textit{None}) because no sufficient condition for any class is met and the decision is rejected. The third example shows a reject based on the category \textit{Both}. In this case two \textit{sufficient} attributes for different classes are detected. Irrespective of whether the predicted class is in the set of all possible classes $\hat{Y}$, at least one other class is also supported. This contradiction leads to a rejection of the decision.

%% file: conclusion.tex
\section{Conclusion}
We have presented an analysis of different dependency decompositions for explainable models and a method to verify the decision of the base-network that performs the intrinsic task. The results have shown that additional attributes can be used to support, verify, and explain the predictions of the network with an option of rejecting the decision. Furthermore, with the right dependencies, the usage of explaining attributes even lead to an increase in accuracy. The thesis of \cite{Turner2015} could be confirmed for some of our models. Although the accuracy of model \miacd \ is slightly below the reference model \mref, the other models show a comparable or increased accuracy. From the increased performance of the models \mcdda \ and \mcdia \ it can be concluded that there are dependencies between the attributes. To obtain reliable classifications with decent explanation it is recommended to use models where the classifier can benefit from an attribute dependency and still has a low rejection rate (e.g. \mcdia). 
Adding the reject option for ambiguous inputs leads to an increase in accuracy for all of our models. In addition the verification process delivers a justification for performing a reject. 
  
As the attributes were gathered supervised, they may not cover the best representation possible from the networks point of view, but they are guaranteed to be understandable by a human. Unsupervised explanation generation systems \cite{Hendricks2016,Vinyals2015} may have the problem to generate explanations that are not directly interpretable. However, the introduced dependency decomposition can also be used or combined with unsupervised procedures (e.g. \cite{Huang2016}). 

%% file: egpaper_final.bbl
\begin{thebibliography}{10}\itemsep=-1pt

\bibitem{Akhtar2017}
N.~Akhtar, J.~Liu, and A.~Mian.
\newblock Defense against universal adversarial perturbations.
\newblock {\em arXiv preprint arXiv:1711.05929}, 2017.

\bibitem{Alsallakh2018}
B.~Alsallakh, A.~Jourabloo, M.~Ye, X.~Liu, and L.~Ren.
\newblock Do convolutional neural networks learn class hierarchy?
\newblock {\em TVCG}, 24(1):152--162, 2018.

\bibitem{Bach2015}
S.~Bach, A.~Binder, G.~Montavon, F.~Klauschen, K.-R. M{\"u}ller, and W.~Samek.
\newblock On pixel-wise explanations for non-linear classifier decisions by
  layer-wise relevance propagation.
\newblock {\em PloS one}, 10(7):e0130140, 2015.

\bibitem{Barber2012}
D.~Barber.
\newblock {\em Bayesian Reasoning and Machine Learning}.
\newblock Cambridge University Press, New York, NY, USA, 2012.

\bibitem{Belnap1977}
N.~D. Belnap.
\newblock {\em A Useful Four-Valued Logic}, pages 5--37.
\newblock Springer Netherlands, Dordrecht, 1977.

\bibitem{Co2019}
K.~T. Co, L.~Mu{\~n}oz-Gonz{\'a}lez, and E.~C. Lupu.
\newblock Procedural noise adversarial examples for black-box attacks on deep
  neural networks.
\newblock {\em arXiv preprint arXiv:1810.00470}, 2019.

\bibitem{Dumitru2009}
D.~Erhan, Y.~Bengio, A.~Courville, and P.~Vincent.
\newblock Visualizing higher-layer features of a deep network.
\newblock Technical report, Univeristé de Montréal, 2009.

\bibitem{Fong2017}
R.~C. Fong and A.~Vedaldi.
\newblock Interpretable explanations of black boxes by meaningful perturbation.
\newblock In {\em ICCV}, pages 3449--3457, 2017.

\bibitem{Goodfellow2015}
I.~J. Goodfellow, J.~Shlens, and C.~Szegedy.
\newblock Explaining and harnessing adversarial examples.
\newblock {\em ICLR}, abs/1412.6572, 2015.

\bibitem{Guo2017}
C.~Guo, G.~Pleiss, Y.~Sun, and K.~Q. Weinberger.
\newblock On calibration of modern neural networks.
\newblock In {\em ICML}, 2017.

\bibitem{Hand2017}
E.~M. Hand and R.~Chellappa.
\newblock Attributes for improved attributes: A multi-task network utilizing
  implicit and explicit relationships for facial attribute classification.
\newblock In {\em AAAI}, 2017.

\bibitem{He2017}
X.~{He} and Y.~{Peng}.
\newblock Fine-grained visual-textual representation learning.
\newblock {\em TCSVT}, pages 1--1, 2019.

\bibitem{Hendricks2016}
L.~A. Hendricks, Z.~Akata, M.~Rohrbach, J.~Donahue, B.~Schiele, and T.~Darrell.
\newblock Generating visual explanations.
\newblock In {\em ECCV}, 2016.

\bibitem{Huang2016}
C.~Huang, C.~C. Loy, and X.~Tang.
\newblock Unsupervised learning of discriminative attributes and visual
  representations.
\newblock {\em CVPR}, pages 5175--5184, 2016.

\bibitem{Hubel1962}
D.~H. Hubel and T.~N. Wiesel.
\newblock Receptive fields, binocular interaction and functional architecture
  in the cat's visual cortex.
\newblock {\em The Journal of physiology}, 160(1):106--154, 1962.

\bibitem{Gal2017}
A.~Kendall and Y.~Gal.
\newblock {What Uncertainties Do We Need in Bayesian Deep Learning for Computer
  Vision?}
\newblock In {\em NIPS}, 2017.

\bibitem{Kindermans2017}
P.-J. Kindermans, S.~Hooker, J.~Adebayo, M.~Alber, K.~T. Sch{\"u}tt,
  S.~D{\"a}hne, D.~Erhan, and B.~Kim.
\newblock The (un)reliability of saliency methods.
\newblock {\em NIPS}, 2017.

\bibitem{Kindermans2017a}
P.-J. {Kindermans}, K.~T. {Sch{\"u}tt}, M.~{Alber}, K.-R. {M{\"u}ller},
  D.~{Erhan}, B.~{Kim}, and S.~{D{\"a}hne}.
\newblock {Learning how to explain neural networks: PatternNet and
  PatternAttribution}.
\newblock {\em arXiv e-prints}, page arXiv:1705.05598, May 2017.

\bibitem{Krizhevsky2012}
A.~Krizhevsky, I.~Sutskever, and G.~E. Hinton.
\newblock Imagenet classification with deep convolutional neural networks.
\newblock In {\em NIPS}, pages 1097--1105, 2012.

\bibitem{Kurakin2017}
A.~Kurakin, I.~J. Goodfellow, and S.~Bengio.
\newblock Adversarial examples in the physical world.
\newblock {\em ICLR}, abs/1607.02533, 2017.

\bibitem{Lapuschkin2016}
S.~Lapuschkin, A.~Binder, G.~Montavon, K.-R. M{{{\"u}}}ller, and W.~Samek.
\newblock The lrp toolbox for artificial neural networks.
\newblock {\em Journal of Machine Learning Research}, 17(114):1--5, 2016.

\bibitem{Liao2018}
F.~Liao, M.~Liang, Y.~Dong, T.~Pang, J.~Zhu, and X.~Hu.
\newblock Defense against adversarial attacks using high-level representation
  guided denoiser.
\newblock In {\em CVPR}, pages 1778--1787, 2018.

\bibitem{Lindh2018}
A.~Lindh, R.~J. Ross, A.~Mahalunkar, G.~Salton, and J.~D. Kelleher.
\newblock Generating diverse and meaningful captions.
\newblock In {\em ICANN}, pages 176--187. Springer, 2018.

\bibitem{Nguyen2017}
A.~Nguyen, J.~Clune, Y.~Bengio, A.~Dosovitskiy, and J.~Yosinski.
\newblock Plug \& play generative networks: Conditional iterative generation of
  images in latent space.
\newblock In {\em CVPR}, pages 3510--3520, July 2017.

\bibitem{Nguyen2014}
A.~Nguyen, J.~Yosinski, and J.~Clune.
\newblock Deep neural networks are easily fooled: High confidence predictions
  for unrecognizable images.
\newblock In {\em CVPR}, 2014.

\bibitem{olah2017feature}
C.~Olah, A.~Mordvintsev, and L.~Schubert.
\newblock Feature visualization.
\newblock {\em Distill}, 2017.
\newblock https://distill.pub/2017/feature-visualization.

\bibitem{Peng2018}
X.~Peng, Z.~Tang, F.~Yang, R.~S. Feris, and D.~Metaxas.
\newblock Jointly optimize data augmentation and network training: Adversarial
  data augmentation in human pose estimation.
\newblock In {\em CVPR}, pages 2226--2234, 2018.

\bibitem{Ribeiro2016}
M.~T. Ribeiro, S.~Singh, and C.~Guestrin.
\newblock "why should i trust you?": Explaining the predictions of any
  classifier.
\newblock In {\em HLT-NAACL Demos}, 2016.

\bibitem{Riesenhuber1999}
M.~Riesenhuber and T.~Poggio.
\newblock Hierarchical models of object recognition in cortex.
\newblock {\em Nature neuroscience}, 2(11):1019, 1999.

\bibitem{Selvaraju2017}
R.~R. Selvaraju, M.~Cogswell, A.~Das, R.~Vedantam, D.~Parikh, and D.~Batra.
\newblock Grad-cam: Visual explanations from deep networks via gradient-based
  localization.
\newblock In {\em ICCV}, pages 618--626, 2017.

\bibitem{Simonyan2013}
K.~Simonyan, A.~Vedaldi, and A.~Zisserman.
\newblock Deep inside convolutional networks: Visualising image classification
  models and saliency maps, 2013.

\bibitem{Springenberg2014}
J.~T. Springenberg, A.~Dosovitskiy, T.~Brox, and M.~A. Riedmiller.
\newblock Striving for simplicity: The all convolutional net.
\newblock {\em ICLR}, abs/1412.6806, 2015.

\bibitem{Stallkamp2012}
J.~Stallkamp, M.~Schlipsing, J.~Salmen, and C.~Igel.
\newblock Man vs. computer: Benchmarking machine learning algorithms for
  traffic sign recognition.
\newblock {\em Neural networks : the official journal of the International
  Neural Network Society}, 32:323--32, 2012.

\bibitem{Sundararajan2017}
M.~Sundararajan, A.~Taly, and Q.~Yan.
\newblock Axiomatic attribution for deep networks, 2017.

\bibitem{Thelisson2017}
E.~Thelisson, K.~Padh, and L.~E. Celis.
\newblock Regulatory mechanisms and algorithms towards trust in ai/ml.
\newblock In {\em IJCAI}, 2017.

\bibitem{Turner2015}
R.~Turner.
\newblock A model explanation system.
\newblock In {\em MLSP}, 2016.

\bibitem{Vinyals2015}
O.~Vinyals, A.~Toshev, S.~Bengio, and D.~Erhan.
\newblock Show and tell: A neural image caption generator.
\newblock {\em CVPR}, pages 3156--3164, 2015.

\bibitem{Xu2015}
K.~Xu, J.~Ba, R.~Kiros, K.~Cho, A.~C. Courville, R.~Salakhutdinov, R.~S. Zemel,
  and Y.~Bengio.
\newblock Show, attend and tell: Neural image caption generation with visual
  attention.
\newblock In {\em ICML}, 2015.

\bibitem{Xu2018}
Y.~Xu, L.~Qin, X.~Liu, J.~Xie, and S.-C. Zhu.
\newblock A causal and-or graph model for visibility fluent reasoning in
  tracking interacting objects.
\newblock In {\em CVPR}, pages 2178--2187, 2018.

\bibitem{Zeiler2014}
M.~D. Zeiler and R.~Fergus.
\newblock Visualizing and understanding convolutional networks.
\newblock In D.~Fleet, T.~Pajdla, B.~Schiele, and T.~Tuytelaars, editors, {\em
  ECCV}, pages 818--833, Cham, 2014. Springer International Publishing.

\bibitem{Zeiler2011}
M.~D. Zeiler, G.~W. Taylor, and R.~Fergus.
\newblock Adaptive deconvolutional networks for mid and high level feature
  learning.
\newblock {\em ICCV}, pages 2018--2025, 2011.

\bibitem{Zhang2018}
Z.~Zhang, C.~Xie, J.~Wang, L.~Xie, and A.~L. Yuille.
\newblock Deepvoting: A robust and explainable deep network for semantic part
  detection under partial occlusion.
\newblock In {\em CVPR}, pages 1372--1380, 2018.

\bibitem{Zhou2014}
B.~Zhou, A.~Khosla, {\`A}.~Lapedriza, A.~Oliva, and A.~Torralba.
\newblock Object detectors emerge in deep scene cnns.
\newblock {\em ICLR}, abs/1412.6856, 2014.

\end{thebibliography}
